\documentclass[conference]{IEEEtran}
\IEEEoverridecommandlockouts

\usepackage{microtype}
\usepackage{graphicx}
\usepackage{enumitem}
\usepackage{subfigure}
\usepackage{algorithm}
\usepackage{booktabs} 
\usepackage[table]{xcolor}
 
\usepackage{hyperref}

\usepackage{amsmath}
\usepackage{amssymb}
\usepackage{mathtools}
\usepackage{amsthm}
\usepackage{caption}

\usepackage[capitalize,noabbrev]{cleveref}

\theoremstyle{plain}

\theoremstyle{definition}

\theoremstyle{remark}

\usepackage[textsize=tiny]{todonotes}

\usepackage{cite}
\usepackage{amsmath,amssymb,amsfonts}
\usepackage{algorithmic}
\usepackage{graphicx}
\usepackage{textcomp}
\usepackage{xcolor}
\def\BibTeX{{\rm B\kern-.05em{\sc i\kern-.025em b}\kern-.08em
    T\kern-.1667em\lower.7ex\hbox{E}\kern-.125emX}}
\begin{document}

\title{Accelerated Smoothing: A Scalable Approach to Randomized Smoothing}

\author{
\IEEEauthorblockN{Devansh Bhardwaj\IEEEauthorrefmark{2}, Kshitiz Kaushik\IEEEauthorrefmark{2}, Sarthak Gupta\IEEEauthorrefmark{2}}
\IEEEauthorblockA{\textit{Indian Institute of Technology, Roorkee, India} \\
\{d\_bhardwaj@ece., k\_kaushik@ee., sarthak\_g@ma.\}iitr.ac.in}
\IEEEauthorblockA{\IEEEauthorrefmark{2}Equal contribution}
}

\maketitle
\begin{abstract}
Randomized smoothing has emerged as a potent certifiable defense against adversarial attacks by employing smoothing noises from specific distributions to ensure the robustness of a smoothed classifier. However, the utilization of Monte Carlo sampling in this process introduces a compute-intensive element, which constrains the practicality of randomized smoothing on a larger scale. To address this limitation, we propose a novel approach that replaces Monte Carlo sampling with the training of a surrogate neural network. Through extensive experimentation in various settings, we demonstrate the efficacy of our approach in approximating the smoothed classifier with remarkable precision. Furthermore, we demonstrate that our approach significantly accelerates the robust radius certification process, providing nearly $600$X improvement in computation time, overcoming the computational bottlenecks associated with traditional randomized smoothing. 
\end{abstract}

\section{Introduction}
Neural networks are known to be vulnerable to perturbed input images, commonly referred to as adversarial attacks. These attacks come in various forms, including but not limited to gradient-based attacks \cite{szegedy2013intriguing}, transfer-based attacks \cite{papernot2016transferability}, and optimization-based attacks \cite{carlini2017towards}. In response to these threats, various empirical defenses have been proposed, such as adversarial training \cite{goodfellow2014explaining} and defensive distillation \cite{papernot2016distillation}. However, these defenses are often susceptible to adaptive attacks and lack theoretical guarantees of their robustness.

Certified defenses have emerged as an alternative approach that aims to provide rigorous theoretical guarantees on the robustness of neural networks against adversarial attacks. Among these, randomized smoothing \cite{lecuyer2019certified}, \cite{cohen2019certified} has gained attention for its ability to scale to higher-dimensional datasets such as Image-Net \cite{deng2009imagenet} and its lack of strong assumptions about the underlying classifier.

Despite its theoretical appeal, there are several challenges that still remain with randomized smoothing such as the well-known "$\ell_\infty$ barrier" or "Curse of Dimensionality" \cite{blum2020random}, or the robustness accuracy trade-off \cite{zhang2019theoretically}, \cite{tsipras2018robustness}. 
\begin{figure}[h]

\begin{flushright}

\includegraphics[width=0.5\textwidth]{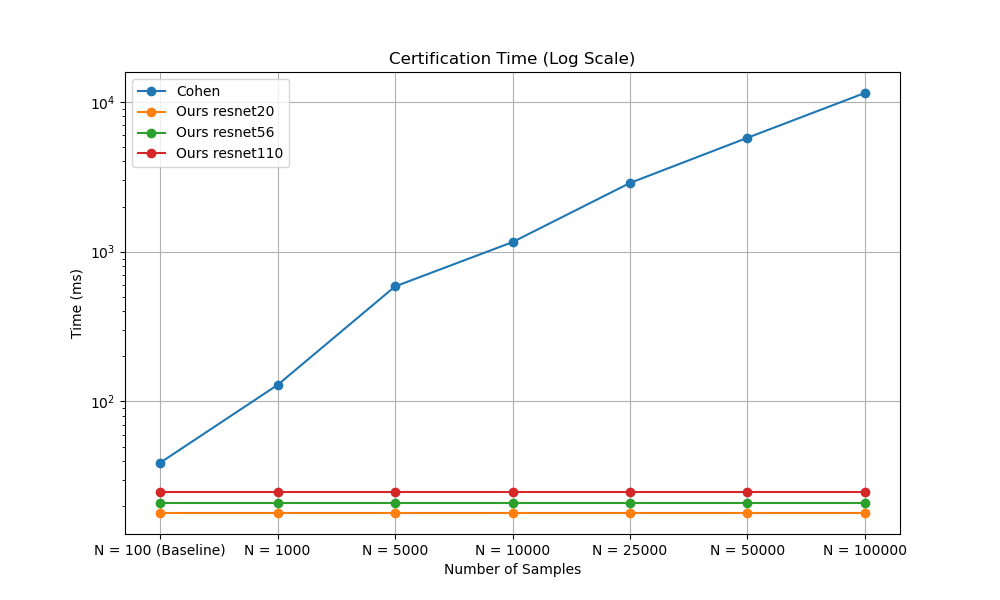}

\caption{Plot of Certification Time vs. No. of Samples. As can be seen, our methodology is $\mathcal{O}(1)$ whereas Neyman-Pearson-based sampling scales linearly, that is, $\mathcal{O}(N)$.}

\label{fig:image}

\end{flushright}

\end{figure}

One crucial challenge that has received less attention is its certification \textit{time complexity}. Randomized smoothing \cite{cohen2019certified}, requires multiple forward passes to obtain a tight lower bound around the top-1 class, leading to increased computational overhead, setting the stage for challenges related to the Average Certified Radius (ACR)-runtime trade-off.

Our proposal, \textbf{Accelerated Smoothing}, aims to tackle this problem by employing a surrogate neural network that has been trained especially for the certification task, thereby avoiding the necessity of making numerous forward passes in randomized smoothing.In addition to significantly reducing inference time, accelerated smoothing closely approximates the original Monte Carlo-based sampling used in randomized smoothing \cite{cohen2019certified}.

Our contributions can be summarized as follows:
\begin{enumerate}[noitemsep,nolistsep,topsep=0pt,partopsep=0pt]
    \item We introduce Accelerated Smoothing, a technique that leverages a surrogate model to accelerate the computation of randomized smoothing-based certified radii.
    \item We show that Accelerated Smoothing could be more efficient in real-world applications since it not only closely imitates the original sampling-based methodology, but also significantly reduces the time required for certification. 
    \item Through extensive empirical evaluation and ablation studies on the CIFAR-10 dataset, we show the effectiveness of Accelerated Smoothing in providing robustness guarantees.
\end{enumerate}

By addressing the time complexity issue of randomized smoothing, Accelerated Smoothing aims to make certified defenses more accessible and practical for deployment in real-world scenarios.

\section{Related Works} 
It has been shown that \textbf{ adversarial attacks} pose a significant threat to the integrity of neural networks \cite{athalye2018obfuscated}, \cite{eykholt2018robust}, \cite{kurakin2018adversarial}. Carefully perturbed samples could be used to fool a machine learning model and cause damage. To overcome this, a plethora of defenses have been proposed which could be broadly divided into empirical and certified defenses. Although \textbf{empirical defenses} have been shown to provide some protection \cite{goodfellow2014explaining}, \cite{attackspeernets}, \cite{buckman2018thermometer}, \cite{guo2017countering}, \cite{dhillon2018stochastic}, \cite{xie2017mitigating}, \cite{song2017pixeldefend} they are often broken later with stronger attacks \cite{tramer2020adaptive}, \cite{athalye2018obfuscated}. On the other hand, \textbf{certified defenses} offer provable robustness guarantees for predictions which can't be broken by new attack methods. We get mathematically proven lower bound on the robust radius around each sample. Various certified defenses have been proposed, such as dual network \cite{dvijotham2018dual}, convex polytope \cite{wong2018provable}, CROWN-IBP \cite{zhang2019towards}, and Lipschitz bounding \cite{cisse2017parseval}. However, these defenses have limitations, including low scalability and strict constraints on the neural network architecture. 

\textbf{Randomized Smoothing} In their seminal work \cite{cohen2019certified}, randomized smoothing was introduced as a defense against $\ell_2$ norm perturbations, outperforming other certified defenses. Subsequent work extended random smoothing to defend against various attacks, including $\ell_0$, $\ell_1$, $\ell_2$, and $\ell_\infty$ norm perturbations, as well as geometric transformations. For example, \cite{levine2020randomized} proposed random ablation to defend against $\ell_0$-norm adversarial attacks, while \cite{yang2020randomized} introduced a uniform Wulff crystal distribution for perturbations of the $\ell_1$ norm. To defend against $\ell_\infty$ norm perturbations, \cite{awasthi2020adversarial} introduced the $\infty \rightarrow 2$ matrix operator for Gaussian smoothing. Randomized smoothing has also been used to defend against adversarial translations by \cite{fischer2020certified}, \cite{li2021tss}. In particular, most of the randomized smoothing works \cite{salman2019provably}, \cite{cohen2019certified}, \cite{zhai2020macer}, \cite{jeong2020consistency}, \cite{yang2020randomized}, \cite{jia2019certified} have achieved superior certified robustness compared to other certified defenses in their respective domains.

\textbf{Issues with Randomized Smoothing} Despite its impressive performance, Randomized Smoothing is not without limitations. \cite{cohen2019certified} primarily focused on $\ell_2$ case attacks and acknowledged that other noise distributions could offer robustness guarantees for different $\ell_p$ attacks. However, it was later discovered that defending against $\ell_p$ attacks with $p > 2$ becomes challenging as it was later observed that for general $\ell_p$ cases with $p>2$, an increase in input dimensions leads to a decreased certified radius. Another concern with randomized smoothing is the trade-off between robustness and accuracy. \cite{cohen2019certified} noted that increasing the variance of the noise can result in a higher certified radius, but this comes at the cost of decreased accuracy in the smoothed model. Several studies have addressed these issues, such as \cite{blum2020random} for the curse of dimensionality and \cite{zhang2019theoretically}, \cite{tsipras2018robustness} for the trade-off between robustness and accuracy. However, one aspect that has received less attention is the time complexity of randomized smoothing.There have been very few works that have built upon the work of \cite{cohen2019certified}. One notable work is Input Specific Robustness\cite{chen2021inputspecific}, which proposes a solution to the problem. In this approach, a mapping is pre-computed for each input, which allows for a trade-off between the Average Certified Radius and the time required. To achieve this, a two-sided Clopper-Pearson \cite{clopper1934use} interval is initially calculated using a smaller sample size. Then, a mapping is generated that shows the relationship between the sample size and the decline in average radius for the input. This mapping demonstrates the trade-off between the number of samples and the average certified radius.

\section{Preliminaries}
\subsection{Randomized Smoothing} 
The basic idea of randomized smoothing (\cite{cohen2019certified}) is to generate a smoothed version of the base classifier $f$. Given an arbitrary base classifier $f(x) : \mathbb{R}^d \rightarrow \mathcal{Y}$ where $\mathcal{Y} = \{1, \ldots, k\}$ is the output space, the smoothed classifier $g(\cdot)$ is defined as:
\begin{equation}
    g(x) := \arg\max_{c \in \mathcal{Y}} \mathbb{P}[f(x + \epsilon) = c], \quad \epsilon \sim \mathcal{N}(0, \sigma^2I^d) \label{eq:smoothed_classifier}
\end{equation}

$g(x)$ returns the most likely predicted label of $f(\cdot)$ when the data is inputted with Gaussian augmentation $\mathcal{N}(x, \sigma^2I^d)$.

\textbf{Theorem}: Let  $f : \mathbb{R}^d \rightarrow \mathcal{Y}$ be any deterministic or random function, and let $\epsilon \sim \mathcal{N} (0, \sigma^2I)$. Let  $g$ be defined as in (\ref{eq:smoothed_classifier}).
Suppose $c_A \in \mathcal{Y}$ and $\underline{p_A}, \overline{p_B} \in [0, 1]$ satisfy:
\begin{equation}
    \label{condition for randomized smoothing}
    \mathbb{P}(f(x + \epsilon) = c_A) \geq \underline{p_A} \geq \overline{p_B} \geq \max_{c \neq c_A} P(f(x + \epsilon) = c)
\end{equation}
where, most probable class $c_A$ is returned with lower bound probability $\underline{p_A}$, and the “runner-up” class is returned with upper bound probability $\overline{p_B}$.

Then $g(x+\delta) = c_A \; \forall \; ||\delta||_2 < R,$ where
\begin{equation}
    \label{certified radius}
    R = \frac{\sigma}{2}(\phi^{-1}(\underline{p_A})-\phi^{-1}(\overline{p_B}))
\end{equation}
where, $\phi^{-1}$ is the inverse of the standard Gaussian CDF.

Moreover, the tight lower bound of $\ell_2$-norm certified radius \cite{cohen2019certified} for the prediction $c_A = g(x)$ is:
\begin{equation}
    \begin{aligned}    
        \sigma\Phi^{-1}(\underline{p_A}) \quad \text{where},  \\
        p_A := \mathbb{P}[f(x + \epsilon) = c_A] > \underline{p_A}, \quad \epsilon \sim \mathcal{N}(0, \sigma^2I^d). 
    \end{aligned}
\end{equation}
 and $\Phi^{-1}$ is the inverse of the standard Gaussian CDF. Since computing the deterministic value of $g(x)$ is impossible because $g(\cdot)$ is built on the random distribution $\mathcal{N}(x, \sigma^2I^d)$, therefore a Monte Carlo sampling-based approach is typically used for estimation of $\underline{p_A}.$
 
\subsection{Monte Carlo based algorithm for Evaluation}
Monte Carlo Sampling is used to predict the class probabilities at $x$ through sampling at different samples of noise. Specifically, to evaluate $g(x)$, $N$ noise samples, that is, $\epsilon_1, \epsilon_2, \epsilon_3, \dots, \epsilon_N$ are drawn from $\mathcal{N}(0, \sigma^2I^d)$ and passed through the base classifier to obtain the predictions $f(x + \epsilon_1), f(x + \epsilon_2), \dots, f(x + \epsilon_N)$. These predictions are then used to return the counts for each class, where the counts for a class $c$ are defined as
\begin{equation}
    class\_counts = \textbf{}\sum_{i=1}^{N} 1[f(x + \epsilon_i) = c]
\end{equation}
Based on these class counts, a two-sided hypothesis test is performed to obtain a lower bound $\underline{p_A}$ on $p_A$. 

\section{Methodology}

    \begin{figure*}
        \centering
        \includegraphics[width=1\linewidth]{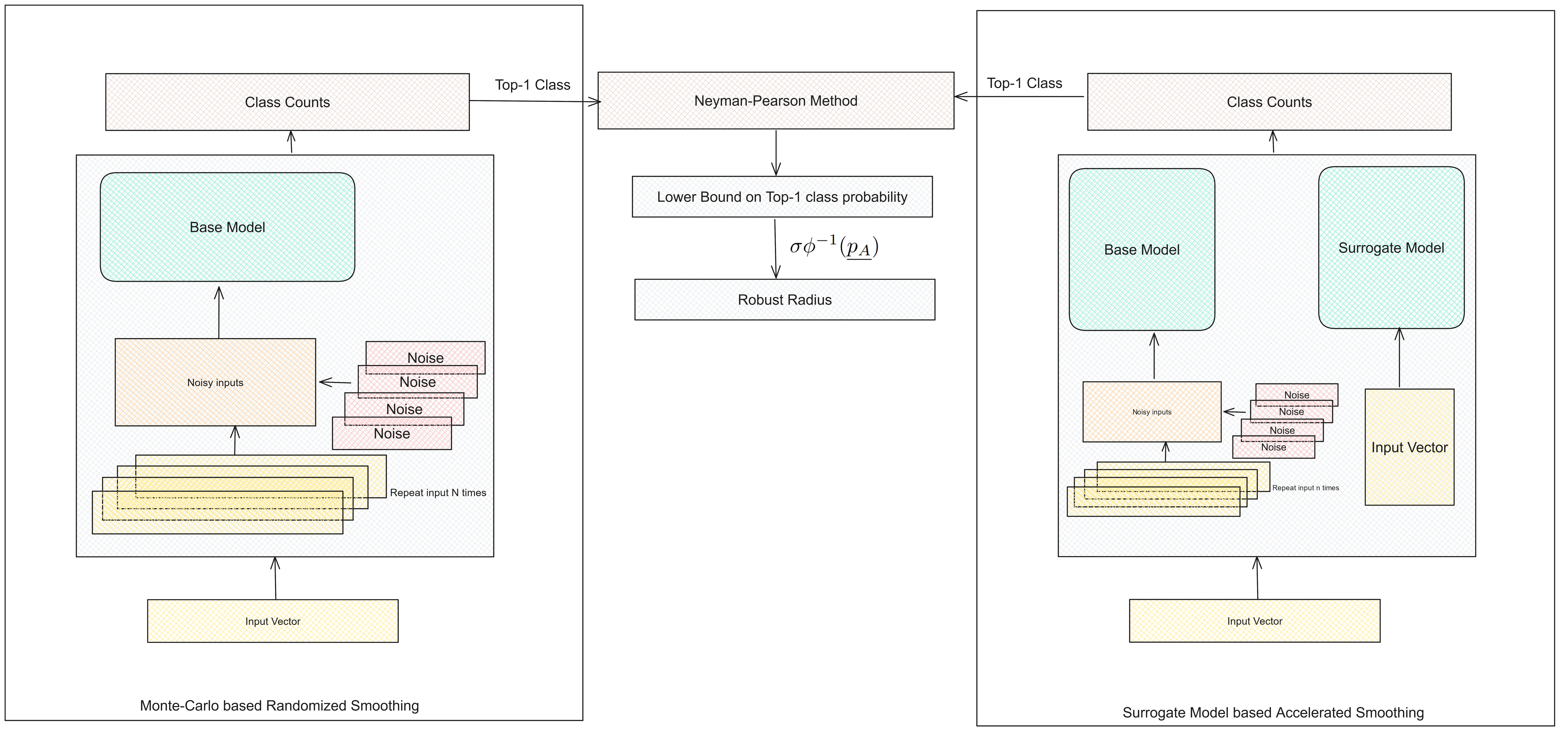}
        \caption{\textbf{\textit{Left:}} Depicts the original Randomized smoothing methodology. \textbf{\textit{Right:}} Illustrates our approach, known as Accelerated Smoothing. In the original methodology, a large number of noise samples are taken, resulting in increased computational cost. However, in our approach, we use a significantly lower value of $N$ for noise samples, solely for the purpose of ensuring that our prediction aligns with that of our surrogate model. The main distinction lies in the surrogate model itself. In the original methodology, the higher value of $N$ is employed to calculate class counts, whereas we utilize a surrogate model to predict the class count based on the input image, thereby significantly reducing the time complexity.}
        \label{fig:enter-label}
    \end{figure*}

\subsection{Motivation}
To obtain a larger certified radius, a large value of $N$ is generally required; for example, Cohen et al. pointed out that for a given noise level $\sigma$ the number of forward passes required for a certified radius of $4\sigma$ is $10^5$. This is costly and will not be feasible in time-constrained settings. 

The smoothed classifier $g$ is argmax on a probability distribution over classes, $\mathcal{Y}$, as indicated in Sections 3.1 and 3.2. The certification algorithm is divided into two components: the first is the Monte Carlo sampling-based probability distribution estimation over classes, which is covered in Section 3.2; the other is the calculation of the lower bound of the top 1 class based on this estimate. The first component is what causes the increased time complexity; our primary goal is to find a significantly less expensive computational solution to Monte Carlo Sampling. 

\subsection{Key Idea}
We propose Accelerated Smoothing as an alternative to the Monte-Carlo based randomized smoothing, specifically, we train a surrogate neural network that aims to predict the class-wise counts that is obtained via Monte Carlo sampling in randomized smoothing. 

Mathematically, we define a surrogate model:
\begin{equation}
  h_{\theta}(x) : R^d \rightarrow C, \text{ where } C = \frac{1}{N}\sum_{i=1}^{N} 1\;[f(x + \epsilon_i) = c]  
\end{equation}
$C $, is the normalized class counts vector obtained from Monte Carlo Sampling. Since we are estimating a probability distribution over classes, hence we use \textit{ JS divergence} as our loss function to train our Surrogate Model. 

The process of training the surrogate model can be summarized as follows:
\begin{itemize}
    \item We sample the class counts for a certain value of the number of noisy samples $N$ for each example in the training set, to obtain a training set $X_{train} = \{x_i, C_i\}_{i=1}^n$, $n$ denotes the number of training samples.
    \item We minimize the loss function, 
\begin{equation}
L = \frac{1}{n}\sum_{i = 1}^nJS(h_\theta(x_i)||C_i)
\end{equation}
where $\theta$ are the parameters of the surrogate model. 
\end{itemize}

\textbf{\textit{Noisy Training Data:}} \newline
It's crucial to remember that the training process for surrogate models could include sampling bias. More specifically, the obtained Monte Carlo samples might be noisy. This means that if we perform multiple samplings and maintain the same input $x$ and number of noisy samples $N$, the resulting class counts list will vary each time.
Consequently, the ground truth value of $C$ is unstable, which complicates the model's ability to train with a single estimate of $C$. 
However, we show in the Appendix \ref{appendix:A} that the variance in the class count is not statistically significant. For every input $x$, it is therefore feasible to use only one sample of $C$.

\textbf{\textit{Generalizability:}}\newline
Our approach is noteworthy because it does not rely on any assumptions about the base classifier or the smoothing distribution. We believe that both of these factors are to some degree reflected in the class counts list obtained for each image. Consequently, our methodology can be readily applied and expanded alongside other advancements in randomized smoothing.

\subsection{Certification via Accelerated Smoothing}
The certification methodology details can be found in \hyperref[alg:Algorithm]{Algorithm}~\ref{alg:Algorithm}. It takes as input a data point $x_i$, the number of noisy samples $N$, a base classifier $f$, a surrogate model $h$, the noise level $\sigma$, the confidence level $\alpha$, and $n_0$, which represents the number of samples used for verifying the top-1 class prediction. The algorithm first uses the base classifier to predict a class $\hat{c}_A$ for the input $x_i$, considering the added noise level $\sigma$ and using $n_0$ samples for verifying the top-1 class. Then it calculates the $counts$ by multiplying the number of noisy samples $N$ by the output of the surrogate model $h(x_i)$. If the top-1 class in the counts matches the predicted class $\hat{c}_A$, the algorithm proceeds to calculate a lower confidence bound $\underline{p_A}$ for the count of the predicted class. If this lower bound is greater than $\frac{1}{2}$, the algorithm returns a certified radius $\sigma\phi^{-1}(\underline{p_A})$, otherwise, it abstains from making a certification decision.
\\
The algorithm outputs the certified radius $R$ if the certification condition is met; otherwise, it abstains from providing certification. 

We'll skip over the specifics of the \verb|PREDICT| and \verb|LowerConfBound| functions here. Please refer to Appendix \ref{appendix:C} for more details on these functions. These functions are straight adaptations of \cite{cohen2019certified}'s certify algorithm.

\textbf{\textit{We are not eliminating the smoothed classifier:}}\newline It is important to emphasize that we only offer the certified radius in cases where the smoothed classifier's predicted top 1 class matches the predicted top 1 class. Moreover, it would indicate that we have trained a surrogate model to completely eliminate the smoothed classifier if this requirement is not established. Our goal is not to use the surrogate model for prediction, but only for robust radius certification. The smoothed classifier won't be entirely eliminated; it will still be utilized in the usual way for predictions. 

\begin{algorithm}[tb]
\caption{Pseudocode for Certification}

\label{alg:Algorithm}

\begin{algorithmic}

\STATE {\bfseries Input:} Data $x_i$, Number of noisy samples $N$, base classifier $f$, surrogate model $h$, Noise Level $\sigma$, Confidence level $\alpha$, $n_0$ no. of samples for verifying top1 class

\STATE $\hat{c}_A = \text{PREDICT}(f,x,\sigma,\alpha,n_0)$

\STATE $counts = N \cdot h(x)$

\STATE $c_0 = \text{top 1 class in counts}$

\IF{$c_0 == \hat{c}_A$}

\STATE $\underline{p_A} = \text{LowerConfBound}(counts[c_0], N, 1-\alpha)$

\IF{$\underline{p_A} > 1/2$}

\STATE \textbf{return} radius $\sigma \cdot \phi^{-1}(\underline{p_A})$

\ELSE

\STATE \textbf{ABSTAIN}

\ENDIF

\ELSE

\STATE \textbf{ABSTAIN}

\ENDIF

\STATE {\bfseries Output:} Certified Radius $R$

\end{algorithmic}

\end{algorithm}

\section{Experiments}
We now share the steps involved in creating the dataset using the Monte Carlo sampling technique, where $N$ represents the total number of noisy samples. Additionally, we share the steps involved in training the surrogate model.
We then present our surrogate model's certification results and contrast them with a \textit{baseline}.
To demonstrate that our model performs significantly better than the results from a smaller number of samples (while maintaining a similar certification time).
With $N=100$, we considered \textit{baseline} to be a simple randomized smoothing certification.

\subsection{Implementation Details}
\textbf{Dataset Sampling.}  $f(x)$ must consistently classify $\mathcal{N} (x, \sigma^2I)$ as $c$ in order for $g(x)$ from \eqref{eq:smoothed_classifier} to classify the labeled example $(x, c)$ correctly and robustly. The Gaussian distribution $\mathcal{N} (x, \sigma^2I)$ places practically no mass near its mode $x$ in higher dimensional spaces. Consequently, the distribution of natural images has virtually disjoint support from the distribution of natural images corrupted by $\mathcal{N} (x, \sigma^2I)$ when $\sigma$ is moderately high.

Therefore, we adopt a pretrained base classifier augmented with Gaussian data manipulation employing the variance $\sigma^2$, as detailed in \cite{cohen2019certified}. The purpose is to generate samples for each class, denoted as $class\_counts \: \forall \: c \in \mathcal{Y}$, with $\sigma$ taking values from the set $\{0.12, 0.25, 0.50, 1.00\}$.

In the context of a given input $x$, we empirically generated a set of $N$ perturbed samples from the Standard Gaussian distribution, characterized by $\epsilon_i \sim \mathcal{N}(0, \sigma^2I)$. The resultant variable $class\_counts$ denotes a frequency distribution corresponding to the predictions made by the model on perturbed inputs, specifically $f(x + \epsilon_i)$.
In order to represent a suitable probability distribution over $class\_counts$, a normalization procedure is applied. This normalization process ensures that the counts are adjusted to represent a probability distribution over the classes $c \in \mathcal{Y}.$

\textbf{Surrogate Model Training.} 
To learn the mapping from $R^d \rightarrow C$, we employ an architecture based on the ResNet framework. 
To systematically investigate the sensitivity of the results to the escalating complexity of the model, we conducted experiments utilizing variations of the ResNet architecture, specifically {resnet20}, {resnet56}, and {resnet110}, with \textit{JS divergence} as the loss function. The surrogate model underwent training for a duration of $200$ epochs, employing a batch size of $128$ and a learning rate of $1e^{-3}$. The optimization process utilized the {Adam} optimizer with \textit{betas} set to $(0.5, 0.999)$. Additionally, the learning rate scheduling was accomplished through the StepLR method, where the step size was set to $20$, and the decay factor (\textit{gamma}) was configured as $0.5$.

\textbf{Computation Time and Resources.} Our model was trained on individual V100 GPUs, each with 16GB of VRAM. The training time for different variations of the model was as follows: resnet 20 took 900 seconds, resnet 56 took approximately 3600 seconds, and resnet 110 took 4200 seconds. The majority of the time was spent on data sampling, specifically on sampling class counts from the complete dataset of N = 100000, which took around 180-190 hours.

\begin{table*}[t]
\centering
\captionsetup{justification=centering}
    \renewcommand{\arraystretch}{1} 
    \begin{tabular}{|c||c||c|c|c|c|c|c||c|}
    \hline
    $\sigma$ & Method & $r=0.25$ & $r=0.5$ & $r=0.75$ & $r=1$ & $r=1.25$ & $r=1.5$ & ACR \\
    \hline\hline
    \rowcolor{gray}
    0.12 & Cohen (N = 100000) & 59.7\% & 0.0\% & 0.0\% & 0.0\% & 0.0\% & 0.0\% & 0.275 \\
    \cline{2-9}
    & Baseline ( N = 100 ) & 0.0\% & 0.0\% & 0.0\% & 0.0\% & 0.0\% & 0.0\% & 0.124 \\
    \cline{2-9}
    & Ours (resnet 20) & 55.5\% & 0.0\% & 0.0\% & 0.0\% & 0.0\% & 0.0\% & 0.244 \\
    \cline{2-9}
    & Ours (resnet 56) & \textbf{56.4\%} & 0.0\% & 0.0\% & 0.0\% & 0.0\% & 0.0\% & 0.248 \\
    \cline{2-9}
    & Ours (resnet 110) & \textbf{56.4\%} & 0.0\% & 0.0\% & 0.0\% & 0.0\% & 0.0\% & \textbf{0.249} \\
    \cline{2-9}
    \hline\hline
    \rowcolor{gray}
    0.25 & Cohen (N = 100000) & 63.2\% & 43.1\% & 25.1\% & 0.0\% & 0.0\% & 0.0\% & 0.435 \\
    \cline{2-9}
    & Baseline ( N = 100 ) & 50.8\% & 0.0\% & 0.0\% & 0.0\% & 0.0\% & 0.0\% & 0.206 \\
    \cline{2-9}
    & Ours (resnet 20) & 58.5\% & \textbf{42.3\%} & \textbf{20.6\%} & 0.0\% & 0.0\% & 0.0\% & \textbf{0.390} \\
    \cline{2-9}
    & Ours (resnet 56) & \textbf{59.1\%} & 40.9\% & 19.8\% & 0.0\% & 0.0\% & 0.0\% & 0.386 \\
    \cline{2-9}
    & Ours (resnet 110) & 58.3\% & 39.5\% & 19.8\% & 0.0\% & 0.0\% & 0.0\% & 0.383 \\
    \cline{2-9}
    \hline\hline
    \rowcolor{gray}
    0.50 & Cohen (N = 100000) & 54.9\% & 42.7\% & 32.4\% & 21.9\% & 14.0\% & 8.0\% & 0.524 \\
    \cline{2-9}
    & Baseline ( N = 100 ) & 44.0\% & 30.8\% & 15.3\% & 0.0\% & 0.0\% & 0.0\% & 0.275 \\
    \cline{2-9}
    & Ours (resnet 20) & 52.0\% & \textbf{43.1\%} & \textbf{33.0\%} & \textbf{22.1\%} & \textbf{12.5\%} & 5.5\% & \textbf{0.494} \\
   \cline{2-9}
    & Ours (resnet 56) & \textbf{52.1\%} & 42.1\% & 31.5\% & 20.6\% & 12.1\% & \textbf{5.7\% }& 0.487 \\
    \cline{2-9}
    & Ours (resnet 110) & 51.9\% & 41.6\% & 30.7\% & 20.1\% & 11.8\% & 5.6\% & 0.483 \\
    \cline{2-9}
    \hline\hline
    \rowcolor{gray}
    1.00 & Cohen (N = 100000) & 44.0\% & 36.2 & 29.5 & 23.2 & 17.9 & 13.7 & 0.541 \\
    \cline{2-9}
    & Baseline ( N = 100 ) & 29.5\% & 23.2\% & 17.5\% & 12.1\% & 5.8\% & 3.6\% & 0.271 \\
    \cline{2-9}
    & Ours (resnet 20) & \textbf{38.8}\% & \textbf{34.0}\% & \textbf{28.6}\% & \textbf{23.0}\% & \textbf{18.1}\% & \textbf{13.4}\% & \textbf{0.499} \\
    \cline{2-9}
    & Ours (resnet 56) & 38.4\% & 33.1\% & 27.5\% & 22.0\% & 17.2\% & 12.9\% & 0.487 \\
    \cline{2-9}
    & Ours (resnet 110) & 38.3\% & 33.0\% & 27.2\% & 21.8\% & 16.8\% & 12.5\% & 0.484 \\
    \cline{2-9}
    \hline
    \end{tabular}
    \caption{Certified Accuracy; our methodology compared with the original Cohen and Baseline with $N = 100000$ and $N=100$ respectively.}
    \label{tab:Certified Accuracy}
\end{table*}

\subsection{Certification Results}
In the context of adversarially robust classification, a noteworthy metric is the \textit{certified test set accuracy} at radius $r$. This metric is defined as the fraction of the test set for which the classifier $g$ produces correct classifications with predictions that are certifiably robust within an $L_2$ ball of radius $r$.
However, in the case where $g$ is a randomized smoothing classifier, precise computation of this quantity becomes infeasible. Hence, we opt to present an \textit{approximate certified test set accuracy} as proposed by \cite{cohen2019certified}. This accuracy is defined as the proportion of the test set for which the CERTIFY algorithm correctly classifies (without abstaining) and certifies robustness with a radius $r \leq R$. 

Our findings are contrasted with a \textbf{baseline}, which is equivalent to a $N=100$ randomized smoothing certification.
We demonstrate that our methodology yields a trend that is strikingly similar to the original Cohen results, while the baseline falls short of this goal as the certification accuracy rapidly fall to 0\%.
This is further demonstrated by the fact that, while our approach only experiences an average decline of 5–10\%, the ACR decreases by more than 50\% on the baseline. 
This illustrates that although a fewer sample approach can be used to reduce the certification time, the quality of the results will be highly compromised; on the other hand, our method provides the best of both worlds, reducing the cost significantly while maintaining the expected trend.



\subsection{Time Complexity Analysis}
The Randomized Smoothing Methodology, exhibits a time complexity that scales linearly with the number of noise samples ($N$). Figure \ref{fig:image}, depicts that Cohen's methodology shows an increasing computational demand as $N$ grows, with execution times ranging from $39 ms$ to $11462 ms$. In contrast, our proposed method, implemented in resnet20, resnet56, and resnet110, consistently outperforms \cite{cohen2019certified} approach, maintaining a constant execution time across all tested values of $N$.

The constant-order time complexity of our model is ascribed to its acquired learning of the class count probability distribution. Crucially, our method's scalability and efficiency are further highlighted by its ability to compute results in a single forward pass. This allows us to achieve an average computation time reduction of approximately 600\% for $N=100000$ samples, when compared to Cohen's methodology, resulting in a significant drop in computational time, instantiating scalability and expansion of this method even further.

\begin{table*}[t]
\centering
    \renewcommand{\arraystretch}{1} 
    \begin{tabular}{|c|ccccccc|}
    \hline
    Methodology & N=100 & N=1000 & N=5000 & N=10000 & N=25000 & N=50000 & N=100000 \\
    \hline\hline
    \rowcolor{gray}
    Cohen & 39 ms & 129 ms & 586 ms & 1158 ms & 2874 ms & 5756 ms & 11462 ms \\
    resnet20 & 18 ms & 18 ms & 18 ms & 18 ms & 18 ms & 18 ms & 18 ms \\
    resnet56 & 21 ms & 21 ms & 21 ms & 21 ms & 21 ms & 21 ms & 21 ms \\
    resnet110 & 25 ms & 25 ms & 25 ms & 25 ms & 25 ms & 25 ms & 25 ms \\
    \hline
    \end{tabular}
    \caption{Certification Time }
    \label{tab:Time_complexity}
\end{table*}

For our experiments, we established $N_0=100$ as the baseline for certifying the radius, as elaborated in Section 4.0. This choice yielded results closely aligned with the original method.

\begin{table*}[t]
  \centering
  \captionsetup{justification=centering}
  \renewcommand{\arraystretch}{1} 
  \begin{tabular}{|c|c|c|c|c|c|}
    \hline
    & & \multicolumn{2}{c}{Underestimation} & \multicolumn{2}{c|}{Overestimation} \\
    \hline
    $\sigma$ & Model & Percentage Error & Ground ACR & Percentage Error & Ground ACR \\
    \hline
    0.25 & resnet20 & 16.8\% & 0.785 & 32.7\% & 0.533 \\
    & resnet56 & 17.0\% & 0.772 & 29.3\% & 0.532 \\
    & resnet110 & 16.8\% & 0.767 & 27.0\% & 0.539 \\
    \hline
    0.5 & resnet20 & 19.5\% & 1.167 & 36.9\% & 0.762 \\
    & resnet56 & 19.0\% & 1.124 & 29.9\% & 0.785 \\
    & resnet110 & 18.0\% & 1.107 & 27.1\% & 0.786 \\
    \hline
  \end{tabular}
  \label{tab:estimation}
  \caption{Under and Overestimation of certified radius, for $\sigma = 0.25, \; \sigma = 0.5$ whenever $r \geq 0.25$.
  where, \textbf{Percentage Error} represents the average percentage error between the $r_{sampling}$ and $r_{predicted}$.}
\end{table*}

\subsection{Error Analysis}
The sampling method yields a certified radius that approximates the lower bound probability $\underline{p_A}$. Therefore, we have $r_{sampling} \leq r_{theoretical}$, where $r_{sampling}$ and $r_{theoretical}$ represent the radii obtained through Monte Carlo sampling and the theoretical radius, respectively. We utilize a surrogate model to learn the probability distribution and predict $\underline{p_A}$. 

One drawback of using a surrogate model is that it tends to underestimate or overestimate $r_{predicted}$. This discrepancy is evident in Table \ref{tab:Certified Accuracy}, where the certified accuracy results of Cohen and our approach differ. The surrogate model should not deviate significantly from $r_{sampling}$, as the sampling method already underestimates the radius compared to the theoretical radius. Overestimation is also problematic, as stated by Cohen: "If (\ref{condition for randomized smoothing}) is all that is known about $f$, then it is impossible to certify an $\ell_2$ ball with radius larger than $R$." Therefore, minimizing overestimation is crucial. 

However, having an overestimated radius does not necessarily mean that our predicted radius is greater than $r_{theoretical}$, as increasing the value of $N$ can lead to an increase in $r_{sampling}$. 

In this section, we provide an analysis of both underestimation and overestimation. The results are summarized in Table 3, and more detailed table can be found in the Appendix \ref{appendix:B}.\\ 
\textbf{Note} that we only considered underestimation and overestimation analysis when both $r_{predicted}$ and $r_{sampling}$ are greater than $0.25$.  In general, we observed that our model tends to underestimate more frequently than overestimate. Interestingly, both underestimation and overestimation have a similar absolute value of mean error, but the mean ground truth ACR is higher for underestimation compared to overestimation. This implies that underestimation occurs for higher values of the certified radius, while overestimation occurs for lower values. Consequently, the percentage error in overestimation is greater than the percentage error in underestimation.

\begin{figure}

\centering

\includegraphics[width=1\linewidth]{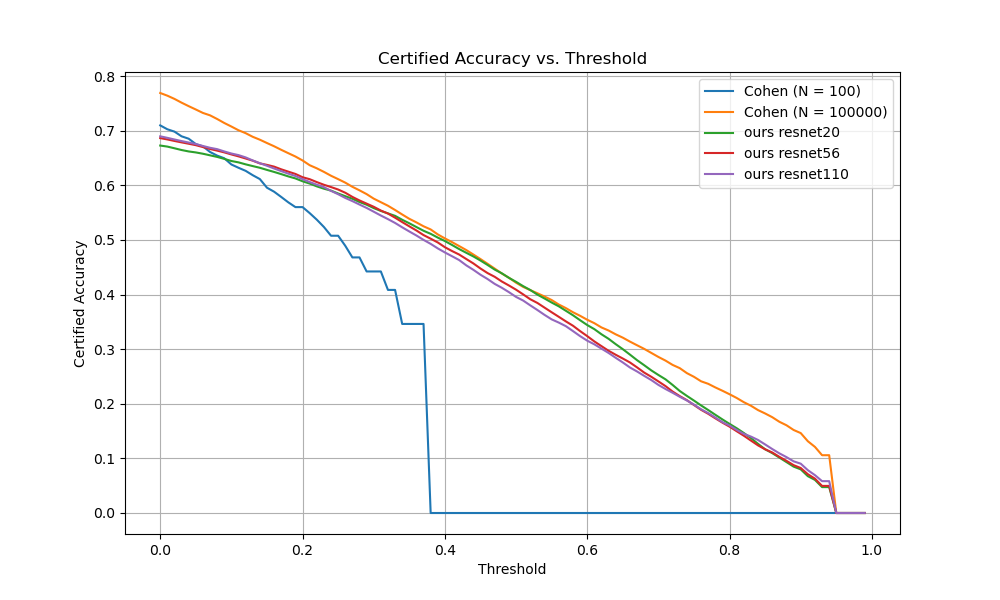}

\includegraphics[width=1\linewidth]{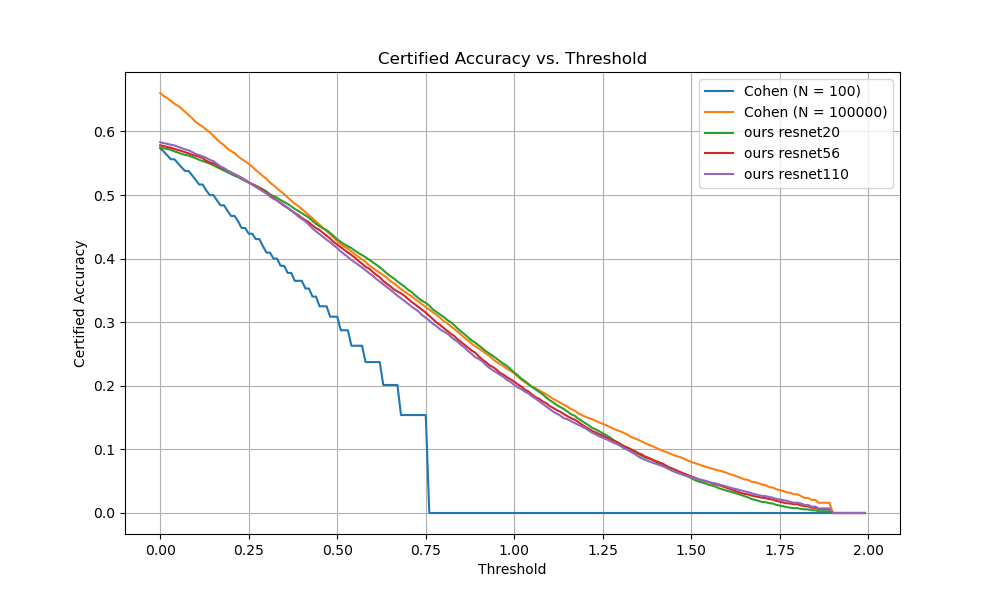}
\caption{Comparison of the certified radius obtained using our method and \cite{cohen2019certified}'s methodology for two different sample sizes, N = 100 and N = 100000. The \textbf{top} section displays the results for $\sigma = 0.25$, while the \textbf{bottom} section displays the results for $\sigma = 0.5$. It can be observed that our methodology allows us to accurately estimate the results for N = 100000.}

\label{fig:enter-label}

\end{figure}

\section{Limitations and Future Works}
Although we have shown good empirical results with significantly reduced time complexity with Accelerated Smoothing. There are still limitations to our approach. The main issue with our methodology lies in the lack of mathematical understanding. Even though the Monte-Carlo sampling method is being replaced by a Surrogate Model, there is still a requirement for a stronger mathematical framework to support the estimates made by the surrogate model.

Another downside of our approach is the dataset sampling time, although during inference the computational costs are reduced significantly, but for sampling the training set requires a lot of time which adds to computational costs during training of the model.

Coming to future works, An interesting direction is to see how our approach scales to other types of randomized smoothing approaches and different training methodologies, since our method is model agnostic hence we believe our methodology should be easily applicable to them. This also implies that other issues such as curse of dimensionality and robustness accuracy trade off might affect our model as well. 

One big issue we face is that our model tends to give similar outputs for the same input with no information about it's certainty, leading to estimation problems, as explained in Section 5.4. To improve this, we suggest training the model to recognize the uncertainty in its outputs. This can be used in deciding how many samples ($N_0$) are needed for radius certification. Many fields want this uncertainty info from Neural Networks. We think using ideas from Bayesian Neural Networks (BNNs) can make our method more reliable.

Other important improvements include trying different sampling methods like \cite{li2022double} and using better training techniques for smoothened model ,\cite{jeong2021smoothmix}, \cite{jeong2020consistency}.

In conclusion, a novel approach to randomized smoothing is presented in this work. We seek to improve our method's effectiveness and practicality for a range of applications by tackling the problem of identical outputs and integrating uncertainty estimation, BNNs, and enhanced techniques as we carry forward our work.

\section{Conclusion}

To summarize, we have introduced a new method called Accelerated Smoothing for verifying the robustness of deep learning models. This method involves training a surrogate neural network to predict the counts obtained from Monte Carlo sampling, where the counts represent the frequency of each class in the perturbed samples. The surrogate model is trained using the Jensen-Shannon divergence as the loss function, aiming to minimize the difference between the predicted counts and the actual counts obtained from the original classifier with added noise. Our experiments demonstrate that the Accelerated Smoothing approach can provide competitive certified radii for deep learning models, while also offering computational advantages compared to traditional randomized smoothing methods. By using a surrogate model to approximate the behavior of the original classifier with added noise, we can efficiently estimate the model's robustness and certify its predictions against adversarial attacks. Overall, Accelerated Smoothing shows promise in improving the scalability and efficiency of certified defenses in deep learning, thereby enhancing the security and reliability of machine learning systems in real-world applications.

\section{Impact Statement}
The potential social implications of our research are significant, as the reliability of machine learning models is crucial for their use in safety-critical systems such as autonomous vehicles, medical diagnosis, and financial forecasting. Our approach can enhance the certification process, making it much faster. This, in turn, can lead to increased safety and dependability in various fields. Although our work aligns with the broader objective of improving the reliability of machine learning systems, it is essential to acknowledge that the implementation of AI technologies should be accompanied by careful consideration of their potential social impact. This includes addressing concerns related to fairness, transparency, and accountability. By offering a method to speed up the certification process of AI systems, our research contributes to the ongoing discussion on the responsible development and deployment of AI technologies.

\newpage
\onecolumn

\section{Appendix}

\subsection{Variance of Sampled dataset}
\label{appendix:A}
In order to demonstrate that observing the training dataset only once through sampling is sufficient, we conducted an ablation study to ascertain the effect of multiple samplings on the dataset in terms of varying $class\_counts$ and measured the sampled dataset's variance.
\\
Over the course of the study, we computed the mean to determine the average variance per-class by sampling $50$ times over $50$ images, each with $N=100000$ noisy samples.
Moreover, the mean variance was within the range of 5\% of all samples, or $N=100000$, despite the variance increase with corresponding increase in $\sigma$.

\begin{table*}[htp]
\centering
    \renewcommand{\arraystretch}{1} 
    \begin{tabular}{|c|c|c|c|c|c|c|c|c|c|c|}
    \hline
    $\sigma$ & \multicolumn{10}{c|}{$c_i \;:\; i^{th}$ class}\\
    \hline\hline
    0.12 & 65.56 & 24.94 & 785.50 & 2060.00 & 1231.00 & 1901.00 & 1204.00 & 849.00 & 671.00 & 83.75\\
    0.25 & 625.5 & 1628 & 1518 & 1779 & 3698 & 2484 & 1222 & 1449 & 998.5 & 1029\\
    0.50 & 2166 & 2860 & 4228 & 3072 & 4392 & 2380 & 2840 & 2240 & 2222 & 2328\\
    1.00 & 2740 & 4228 & 4952 & 5244 & 6236 & 4320 & 4728 & 4240 & 3732 & 3344\\
    \hline
    \end{tabular}
    \caption{Average Sampling Variance (per-class) for $N=100000$.}
    \label{tab:Variance of Sampled dataset (per class)}
\end{table*}

\begin{table*}[htp]
\centering
    \renewcommand{\arraystretch}{1} 
    \begin{tabular}{|c|c|}
    \hline
    $\sigma$ & variance\\
    \hline\hline
    0.12 & 0.88758\% \\
    0.25 & 1.64310\%\\
    0.50 & 2.87280\%\\
    1.00 & 4.37640\%\\
    \hline
    \end{tabular}
    \caption{Average Sampling Variance (\%) for $N = 100000$.}
    \label{tab:Variance of Sampled dataset}
\end{table*}

\subsection{Error Analysis}
\label{appendix:B}
\begin{table*}[htp]
\centering
  
  \caption{The certified radius is either underestimated or overestimated for $\sigma = 0.25$ and $\sigma = 0.5$ when $r \geq 0.25$. The Percentage Error represents the average percentage difference between the sampled radius $r_{sampling}$ and the predicted radius $r_{predicted}$. \textit{Underestimation}\% and \textit{Overestimation}\% indicate the proportion of examples in the entire test set where underestimation and overestimation occurred, respectively.}

\begin{tabular}{lrrrrrr}
\hline
\multicolumn{1}{|l}{} &
  \multicolumn{1}{l}{} &
  \multicolumn{5}{c|}{Underestimation} \\ \hline
\multicolumn{1}{|l|}{sigma} &
  \multicolumn{1}{l|}{Model} &
  \multicolumn{1}{l|}{Percentage Error} &
  \multicolumn{1}{l|}{GT ACR} &
  \multicolumn{1}{l|}{Underestimation \%} &
  \multicolumn{1}{l|}{Mean Error} &
  \multicolumn{1}{l|}{Error Variance} \\ \hline
\multicolumn{1}{|l|}{} &
  \multicolumn{1}{r|}{20} &
  \multicolumn{1}{r|}{16.8} &
  \multicolumn{1}{r|}{0.785} &
  \multicolumn{1}{r|}{33.1} &
  \multicolumn{1}{r|}{0.126} &
  \multicolumn{1}{r|}{0.0142} \\ \hline
\multicolumn{1}{|r|}{0.25} &
  \multicolumn{1}{r|}{56} &
  \multicolumn{1}{r|}{17} &
  \multicolumn{1}{r|}{0.772} &
  \multicolumn{1}{r|}{35.4} &
  \multicolumn{1}{r|}{0.124} &
  \multicolumn{1}{r|}{0.013} \\ \hline
\multicolumn{1}{|l|}{} &
  \multicolumn{1}{r|}{110} &
  \multicolumn{1}{r|}{16.8} &
  \multicolumn{1}{r|}{0.767} &
  \multicolumn{1}{r|}{35.6} &
  \multicolumn{1}{r|}{0.123} &
  \multicolumn{1}{r|}{0.0139} \\ \hline
\multicolumn{1}{|l|}{} &
  \multicolumn{1}{r|}{20} &
  \multicolumn{1}{r|}{19.5} &
  \multicolumn{1}{r|}{1.167} &
  \multicolumn{1}{r|}{26.7} &
  \multicolumn{1}{r|}{0.213} &
  \multicolumn{1}{r|}{0.0301} \\ \hline
\multicolumn{1}{|r|}{0.5} &
  \multicolumn{1}{r|}{56} &
  \multicolumn{1}{r|}{19} &
  \multicolumn{1}{r|}{1.124} &
  \multicolumn{1}{r|}{28.5} &
  \multicolumn{1}{r|}{0.199} &
  \multicolumn{1}{r|}{0.0303} \\ \hline
\multicolumn{1}{|l|}{} &
  \multicolumn{1}{r|}{110} &
  \multicolumn{1}{r|}{18} &
  \multicolumn{1}{r|}{1.107} &
  \multicolumn{1}{r|}{30.1} &
  \multicolumn{1}{r|}{0.185} &
  \multicolumn{1}{r|}{0.0239} \\ \hline
 &
  \multicolumn{1}{l}{} &
  \multicolumn{1}{l}{} &
  \multicolumn{1}{l}{} &
  \multicolumn{1}{l}{} &
  \multicolumn{1}{l}{} &
  \multicolumn{1}{l}{} \\ \hline
\multicolumn{1}{|l}{} &
  \multicolumn{1}{l}{} &
  \multicolumn{5}{c|}{Overestimation} \\ \hline
\multicolumn{1}{|l|}{sigma} &
  \multicolumn{1}{l|}{Model} &
  \multicolumn{1}{l|}{Percentage Error} &
  \multicolumn{1}{l|}{GT ACR} &
  \multicolumn{1}{l|}{Overestimation \%} &
  \multicolumn{1}{l|}{Mean Error} &
  \multicolumn{1}{l|}{Error Variance} \\ \hline
\multicolumn{1}{|l|}{} &
  \multicolumn{1}{r|}{20} &
  \multicolumn{1}{r|}{32.7} &
  \multicolumn{1}{r|}{0.533} &
  \multicolumn{1}{r|}{20.8} &
  \multicolumn{1}{r|}{0.139} &
  \multicolumn{1}{r|}{0.0122} \\ \hline
\multicolumn{1}{|r|}{0.25} &
  \multicolumn{1}{r|}{56} &
  \multicolumn{1}{r|}{29.3} &
  \multicolumn{1}{r|}{0.532} &
  \multicolumn{1}{r|}{19.2} &
  \multicolumn{1}{r|}{0.124} &
  \multicolumn{1}{r|}{0.0104} \\ \hline
\multicolumn{1}{|l|}{} &
  \multicolumn{1}{r|}{110} &
  \multicolumn{1}{r|}{27.0} &
  \multicolumn{1}{r|}{0.539} &
  \multicolumn{1}{r|}{18.7} &
  \multicolumn{1}{r|}{0.116} &
  \multicolumn{1}{r|}{0.0094} \\ \hline
\multicolumn{1}{|l|}{} &
  \multicolumn{1}{r|}{20} &
  \multicolumn{1}{r|}{36.9} &
  \multicolumn{1}{r|}{0.762} &
  \multicolumn{1}{r|}{22.9} &
  \multicolumn{1}{r|}{0.205} &
  \multicolumn{1}{r|}{0.0279} \\ \hline
\multicolumn{1}{|r|}{0.5} &
  \multicolumn{1}{r|}{56} &
  \multicolumn{1}{r|}{29.9} &
  \multicolumn{1}{r|}{0.785} &
  \multicolumn{1}{r|}{21.4} &
  \multicolumn{1}{r|}{0.173} &
  \multicolumn{1}{r|}{0.0218} \\ \hline
\multicolumn{1}{|l|}{} &
  \multicolumn{1}{r|}{110} &
  \multicolumn{1}{r|}{27.1} &
  \multicolumn{1}{r|}{0.786} &
  \multicolumn{1}{r|}{20} &
  \multicolumn{1}{r|}{0.157} &
  \multicolumn{1}{r|}{0.0183} \\ \hline
\end{tabular}
\end{table*}

\subsection{LowerConfBound \& Predict function}
\label{appendix:C}
\textbf{LowerConfBound} ($k$, $n$, $1 \text{-} \alpha$) returns a one-sided $(1 \text{-} \alpha)$ lower confidence interval for the Binomial parameter $p$ given that $k$ $\sim$ Binomial($n, p$). In other words, it returns some number $p$ for which $\underline{p} \leq p$ with probability at least $1 \text{-} \alpha$ over the sampling of $k$ $\sim$ Binomial($n, p$). Following \cite{lecuyer2019certified}, we chose to use the Clopper-Pearson confidence interval, which inverts the Binomial CDF \cite{clopper1934use}. Using 
\verb|statsmodels.stats.proportion.proportion_confint|, this can be implemented as\\
\verb|proportion_confint(k, n, alpha=2*alpha, method="beta")[0]|

\textbf{PREDICT} function given in pseudo-code leverages the hypothesis test given in \cite{Hung & Fithian (2019)} for identifying the top category of a multinomial distribution. \verb|PREDICT| has one tunable hyper-parameter; $\alpha$. When $\alpha$ is small, \verb|PREDICT| abstains frequently but rarely returns the wrong class. When $\alpha$ is large, \verb|PREDICT| usually makes a prediction, but may often return the wrong class.

\begin{thebibliography}{00}

\bibitem{szegedy2013intriguing}
Szegedy, Christian, Wojciech Zaremba, Ilya Sutskever, Joan Bruna, Dumitru Erhan, Ian Goodfellow, and Rob Fergus. "Intriguing properties of neural networks." arXiv preprint arXiv:1312.6199 (2013).

\bibitem{zhang2019theoretically}
Zhang, Hongyang, Yaodong Yu, Jiantao Jiao, Eric Xing, Laurent El Ghaoui, and Michael Jordan. "Theoretically principled trade-off between robustness and accuracy." In International conference on machine learning, pp. 7472-7482. PMLR, 2019.

\bibitem{jeong2020consistency}
Jeong, Jongheon, and Jinwoo Shin. "Consistency regularization for certified robustness of smoothed classifiers." In Advances in Neural Information Processing Systems, vol. 33, pp. 10558-10570. 2020.

\bibitem{jeong2021smoothmix}
Jeong, Jongheon, Sejun Park, Minkyu Kim, Heung-Chang Lee, Do-Guk Kim, and Jinwoo Shin. "SmoothMix: Training Confidence-calibrated Smoothed Classifiers for Certified Robustness." In Advances in Neural Information Processing Systems, vol. 34, pp. 30153-30168. 2021.

\bibitem{tsipras2018robustness}
Tsipras, Dimitris, Shibani Santurkar, Logan Engstrom, Alexander Turner, and Aleksander Madry. "Robustness may be at odds with accuracy." arXiv preprint arXiv:1805.12152 (2018).

\bibitem{blum2020random}
Blum, Avrim, Travis Dick, Naren Manoj, and Hongyang Zhang. "Random smoothing might be unable to certify $l_\infty$ robustness for high-dimensional images." The Journal of Machine Learning Research 21, no. 1 (2020): 8726-8746.

\bibitem{papernot2016transferability}
Papernot, Nicolas, Patrick McDaniel, and Ian Goodfellow. "Transferability in machine learning: from phenomena to black-box attacks using adversarial samples." arXiv preprint arXiv:1605.07277 (2016).

\bibitem{carlini2017towards}
Carlini, Nicholas, and David Wagner. "Towards evaluating the robustness of neural networks." In 2017 ieee symposium on security and privacy (sp), pp. 39-57. IEEE, 2017.

\bibitem{goodfellow2014explaining}
Goodfellow, Ian J, Jonathon Shlens, and Christian Szegedy. "Explaining and harnessing adversarial examples." arXiv preprint arXiv:1412.6572 (2014).

\bibitem{papernot2016distillation}
Papernot, Nicolas, Patrick McDaniel, Xi Wu, Somesh Jha, and Ananthram Swami. "Distillation as a defense to adversarial perturbations against deep neural networks." In 2016 IEEE symposium on security and privacy (SP), pp. 582-597. IEEE, 2016.

\bibitem{lecuyer2019certified}
Lecuyer, Mathias, Vaggelis Atlidakis, Roxana Geambasu, Daniel Hsu, and Suman Jana. "Certified robustness to adversarial examples with differential privacy." In 2019 IEEE symposium on security and privacy (SP), pp. 656-672. IEEE, 2019.

\bibitem{deng2009imagenet}
Deng, Jia, Wei Dong, Richard Socher, Li-Jia Li, Kai Li, and Li Fei-Fei. "Imagenet: A large-scale hierarchical image database." In 2009 IEEE conference on computer vision and pattern recognition, pp. 248-255. IEEE, 2009.

\bibitem{cohen2019certified}
Cohen, Jeremy, Elan Rosenfeld, and Zico Kolter. "Certified adversarial robustness via randomized smoothing." In International conference on machine learning, pp. 1310-1320. PMLR, 2019.

\bibitem{athalye2018obfuscated}
Athalye, Anish, Nicholas Carlini, and David Wagner. "Obfuscated gradients give a false sense of security: Circumventing defenses to adversarial examples." In International conference on machine learning, pp. 274-283. PMLR, 2018.

\bibitem{eykholt2018robust}
Eykholt, Kevin, Ivan Evtimov, Earlence Fernandes, Bo Li, Amir Rahmati, Chaowei Xiao, Atul Prakash et al. "Robust physical-world attacks on deep learning visual classification." In Proceedings of the IEEE conference on computer vision and pattern recognition, pp. 1625-1634. 2018.

\bibitem{kurakin2018adversarial}
Kurakin, Alexey, Ian J Goodfellow, and Samy Bengio. "Adversarial examples in the physical world." In Artificial intelligence safety and security, pp. 99-112. Chapman and Hall/CRC, 2018.

\bibitem{tramer2020adaptive}
Tramer, Florian, Nicholas Carlini, Wieland Brendel, and Aleksander Madry. "On adaptive attacks to adversarial example defenses." In Advances in neural information processing systems, vol. 33, pp. 1633-1645. 2020.

\bibitem{dvijotham2018dual}
Dvijotham, Krishnamurthy, Robert Stanforth, Sven Gowal, Timothy A Mann, Pushmeet Kohli. "A Dual Approach to Scalable Verification of Deep Networks." UAI 1, no. 2 (2018): 3.

\bibitem{wong2018provable}
Wong, Eric, and Zico Kolter. "Provable defenses against adversarial examples via the convex outer adversarial polytope." In International conference on machine learning, pp. 5286-5295. PMLR, 2018.

\bibitem{zhang2019towards}
Zhang, Huan, Hongge Chen, Chaowei Xiao, Sven Gowal, Robert Stanforth, Bo Li, Duane Boning, Cho-Jui Hsieh. "Towards stable and efficient training of verifiably robust neural networks." arXiv preprint arXiv:1906.06316 (2019).

\bibitem{levine2020randomized}
Levine, Alexander, and Soheil Feizi. "(De) Randomized smoothing for certifiable defense against patch attacks." In Advances in Neural Information Processing Systems, vol. 33, pp. 6465-6475. 2020.

\bibitem{yang2020randomized}
Yang, Greg, Tony Duan, J Edward Hu, Hadi Salman, Ilya Razenshteyn, Jerry Li. "Randomized smoothing of all shapes and sizes." In International Conference on Machine Learning, pp. 10693-10705. PMLR, 2020.

\bibitem{awasthi2020adversarial}
Awasthi, Pranjal, Himanshu Jain, Ankit Singh Rawat, Aravindan Vijayaraghavan. "Adversarial robustness via robust low rank representations." In Advances in Neural Information Processing Systems, vol. 33, pp. 11391-11403. 2020.

\bibitem{fischer2020certified}
Fischer, Marc, Maximilian Baader, Martin Vechev. "Certified defense to image transformations via randomized smoothing." In Advances in Neural information processing systems, vol. 33, pp. 8404-8417. 2020.

\bibitem{li2021tss}
Li, Linyi, Maurice Weber, Xiaojun Xu, Luka Rimanic, Bhavya Kailkhura, Tao Xie, Ce Zhang, Bo Li. "Tss: Transformation-specific smoothing for robustness certification." In Proceedings of the 2021 ACM SIGSAC Conference on Computer and Communications Security, pp. 535-557. 2021.

\bibitem{salman2019provably}
Salman, Hadi, Jerry Li, Ilya Razenshteyn, Pengchuan Zhang, Huan Zhang, Sebastien Bubeck, Greg Yang. "Provably robust deep learning via adversarially trained smoothed classifiers." In Advances in Neural Information Processing Systems, vol. 32, 2019.

\bibitem{zhai2020macer}
Zhai, Runtian, Chen Dan, He Di, Zhang Huan, Boqing Gong, Pradeep Ravikumar, Cho-Jui Hsieh, Liwei Wang. "Macer: Attack-free and scalable robust training via maximizing certified radius." arXiv preprint arXiv:2001.02378 (2020).

\bibitem{li2022double}
Li, Linyi, Jiawei Zhang, Tao Xie, Bo Li. "Double sampling randomized smoothing." arXiv preprint arXiv:2206.07912 (2022).

\bibitem{chen2021inputspecific}
Chen, Ruoxin, Jie Li, Junchi Yan, Ping Li, Bin Sheng. "Input-Specific Robustness Certification for Randomized Smoothing." arXiv preprint arXiv:2112.12084 (2021).

\bibitem{guo2017countering}
Guo, Chuan, Mayank Rana, Moustapha Cisse, Laurens Van Der Maaten. "Countering adversarial images using input transformations." arXiv preprint arXiv:1711.00117 (2017).

\bibitem{dhillon2018stochastic}
Dhillon, Guneet S, Kamyar Azizzadenesheli, Zachary C Lipton, Jeremy Bernstein, Jean Kossaifi, Aran Khanna, Anima Anandkumar. "Stochastic activation pruning for robust adversarial defense." arXiv preprint arXiv:1803.01442 (2018).

\bibitem{xie2017mitigating}
Xie, Cihang, Jianyu Wang, Zhishuai Zhang, Zhou Ren, Alan Yuille. "Mitigating adversarial effects through randomization." arXiv preprint arXiv:1711.01991 (2017).

\bibitem{song2017pixeldefend}
Song, Yang, Taesup Kim, Sebastian Nowozin, Stefano Ermon, Nate Kushman. "Pixeldefend: Leveraging generative models to understand and defend against adversarial examples." arXiv preprint arXiv:1710.10766 (2017).

\bibitem{cisse2017parseval}
Cisse, Moustapha, Piotr Bojanowski, Edouard Grave, Yann Dauphin, Nicolas Usunier. "Parseval networks: Improving robustness to adversarial examples." In International conference on machine learning, pp. 854-863. PMLR, 2017.

\bibitem{jia2019certified}
Jia, Jinyuan, Xiaoyu Cao, Binghui Wang, Neil Zhenqiang Gong. "Certified robustness for top-k predictions against adversarial perturbations via randomized smoothing." arXiv preprint arXiv:1912.09899 (2019).

\bibitem{attackspeernets}
PEERNETS: EXPLOITING PEER WISDOM AGAINST ADVERSARIAL ATTACKS.

\bibitem{buckman2018thermometer}
Buckman, Jacob, Aurko Roy, Colin Raffel, Ian Goodfellow. "Thermometer encoding: One hot way to resist adversarial examples." In International conference on learning representations, 2018.

\bibitem{clopper1934use}
Clopper, C. J., Pearson, E. S. "The use of confidence or fiducial limits illustrated in the case of the binomial." Biometrika 26, no. 4 (1934): 404-413.

\end{thebibliography}
\end{document}